\documentclass[12pt]{article}
\usepackage{amsmath}
\usepackage{graphicx,psfrag,epsf}
\usepackage{enumerate}
\usepackage{natbib}
\usepackage{url} 

\usepackage{algorithm2e}
\usepackage{amssymb}
\usepackage{booktabs}
\usepackage{float}
\usepackage{hyperref}

\newcommand{\blind}{0}

\begin{document}

\def\spacingset#1{\renewcommand{\baselinestretch}%
{#1}\small\normalsize} \spacingset{1}


\if0\blind
{
  \title{\bf A Simple and Effective Model-Based Variable Importance Measure}

  \author{
        Brandon M. Greenwell \thanks{The authors gratefully acknowledge \ldots{}} \\
    Wright State University\\
     and \\     Bradley C. Boehmke \\
    University of Cincinnati\\
     and \\     Andrew J. McCarthy \\
    The Perduco Group\\
      }
  \maketitle
} \fi

\if1\blind
{
  \bigskip
  \bigskip
  \bigskip
  \begin{center}
    {\LARGE\bf A Simple and Effective Model-Based Variable Importance Measure}
  \end{center}
  \medskip
} \fi

\bigskip
\begin{abstract}
In the era of ``big data'', it is becoming more of a challenge to not
only build state-of-the-art predictive models, but also gain an
understanding of what's really going on in the data. For example, it is
often of interest to know which, if any, of the predictors in a fitted
model are relatively influential on the predicted outcome. Some modern
algorithms---like random forests and gradient boosted decision
trees---have a natural way of quantifying the importance or relative
influence of each feature. Other algorithms---like naive Bayes
classifiers and support vector machines---are not capable of doing so
and model-free approaches are generally used to measure each predictor's
importance. In this paper, we propose a standardized, model-based
approach to measuring predictor importance across the growing spectrum
of supervised learning algorithms. Our proposed method is illustrated
through both simulated and real data examples. The R code to reproduce
all of the figures in this paper is available in the supplementary
materials.
\end{abstract}

\noindent%
\vfill

\newpage
\spacingset{1.45} 

\section{Introduction}
\label{sec:introduction}

Complex supervised learning algorithms, such as neural networks (NNs)
and support vector machines (SVMs), are more common than ever in
predictive analytics, especially when dealing with large observational
databases that don't adhere to the strict assumptions imposed by
traditional statistical techniques (e.g., multiple linear regression
which typically assumes linearity, homoscedasticity, and normality).
However, it can be challenging to understand the results of such complex
models and explain them to management. Graphical displays such as
variable importance plots (when available) and partial dependence plots
(PDPs) \citep{friedman-2001-greedy} offer a simple solution \citep[see,
for example,][pp.~367--380]{hastie-elements-2009}. PDPs are
low-dimensional graphical renderings of the prediction function
\(\widehat{f}\left(\boldsymbol{x}\right)\) that allow analysts to more
easily understand the estimated relationship between the outcome and
predictors of interest. These plots are especially useful in
interpreting the output from ``black box'' models. While PDPs can be
constructed for any predictor in a fitted model, variable importance
scores are more difficult to define, and when available, their
interpretation often depends on the model fitting algorithm used.

In this paper, we consider a standardized method to computing variable
importance scores using PDPs. There are a number of advantages to using
our approach. First, it offers a standardized procedure to quantifying
variable importance across the growing spectrum of supervised learning
algorithms. For example, while popular statistical learning algorithms
like random forests (RFs) and gradient boosted decision trees (GBMs)
have there own natural way of measuring variable importance, each is
interpreted differently (these are briefly described in Section
\ref{sec:model-based-vi}). Secondly, our method is suitable for use with
any trained supervised learning algorithm, provided predictions on new
data can be obtained. For example, it is often beneficial (from an
accuracy standpoint) to train and tune multiple state-of-the art
predictive models (e.g., multiple RFs, GBMs, and deep learning NNs
(DNNs)) and then combine them into an ensemble called a \emph{super
learner} through a process called \emph{model stacking}. Even if the
base learners can provide there own measures of variable importance,
there is no logical way to combine them to form an overall score for the
super learner. However, since new predictions can be obtained from the
super learner, our proposed variable importance measure is still
applicable (examples are given in Sections
\ref{sec:ensemble}--\ref{sec:automl}). Thirdly, as shown in Section
\ref{sec:interaction}, our proposed method can be modified to quantify
the strength of potential interaction effects. Finally, since our
approach is based on constructing PDPs for all the main effects, the
analyst is forced to also look at the estimated functional relationship
between each feature and the target---which should be done in tandem
with studying the importance of each feature.

\section{Background}
\label{sec:background}

We are often confronted with the task of extracting knowledge from large
databases. For this task we turn to various statistical learning
algorithms which, when tuned correctly, can have state-of-the-art
predictive performance. However, having a model that predicts well is
only solving part of the problem. It is also desirable to extract
information about the relationships uncovered by the learning algorithm.
For instance, we often want to know which predictors, if any, are
important by assigning some type of variable importance score to each
feature. Once a set of influential features has been identified, the
next step is summarizing the functional relationship between each
feature, or subset thereof, and the outcome of interest. However, since
most statistical learning algorithms are ``black box'' models,
extracting this information is not always straightforward. Luckily, some
learning algorithms have a natural way of defining variable importance.

\subsection{Model-based approaches to variable importance}
\label{sec:model-based-vi}

Decision trees probably offer the most natural model-based approach to
quantifying the importance of each feature. In a binary decision tree,
at each node \(t\), a single predictor is used to partition the data
into two homogeneous groups. The chosen predictor is the one that
maximizes some measure of improvement \(\widehat{i}_t\). The relative
importance of predictor \(x\) is the sum of the squared improvements
over all internal nodes of the tree for which \(x\) was chosen as the
partitioning variable; see \citet{classification-breiman-1984} for
details. This idea also extends to ensembles of decision trees, such as
RFs and GBMs. In ensembles, the improvement score for each predictor is
averaged across all the trees in the ensemble. Fortunately, due to the
stabilizing effect of averaging, the improvement-based variable
importance metric is often more reliable in large ensembles
\citep[see][pg. 368]{hastie-elements-2009}. RFs offer an additional
method for computing variable importance scores. The idea is to use the
leftover out-of-bag (OOB) data to construct validation-set errors for
each tree. Then, each predictor is randomly shuffled in the OOB data and
the error is computed again. The idea is that if variable \(x\) is
important, then the validation error will go up when \(x\) is perturbed
in the OOB data. The difference in the two errors is recorded for the
OOB data then averaged across all trees in the forest.

In multiple linear regression, the absolute value of the \(t\)-statistic
is commonly used as a measure of variable importance. The same idea also
extends to generalized linear models (GLMs). Multivariate adaptive
regression splines (MARS), which were introduced in
\citet{friedman-1991-mars}, is an automatic regression technique which
can be seen as a generalization of multiple linear regression and
generalized linear models. In the MARS algorithm, the contribution (or
variable importance score) for each predictor is determined using a
generalized cross-validation (GCV) statistic.

For NNs, two popular methods for constructing variable importance scores
are the Garson algorithm \citep{interpreting-garson-1991}, later
modified by \citet{back-goh-1995}, and the Olden algorithm
\citep{accurate-olden-2004}. For both algorithms, the basis of these
importance scores is the network's connection weights. The Garson
algorithm determines variable importance by identifying all weighted
connections between the nodes of interest. Olden's algorithm, on the
other hand, uses the product of the raw connection weights between each
input and output neuron and sums the product across all hidden neurons.
This has been shown to outperform the Garson method in various
simulations. For DNNs, a similar method due to \citet{data-gedeon-1997}
considers the weights connecting the input features to the first two
hidden layers (for simplicity and speed); but this method can be slow
for large networks.

\subsection{Filter-based approaches to variable importance}

Filter-based approaches, which are described in
\citet{applied-kuhn-2013}, do not make use of the fitted model to
measure variable importance. They also do not take into account the
other predictors in the model.

For regression problems, a popular approach to measuring the variable
importance of a numeric predictor \(x\) is to first fit a flexible
nonparametric model between \(x\) and the target \(Y\); for example, the
locally-weighted polynomial regression (LOWESS) method developed by
\citet{robust-cleveland-1979}. From this fit, a pseudo-\(R^2\) measure
can be obtained from the resulting residuals and used as a measure of
variable importance. For categorical predictors, a different method
based on standard statistical tests (e.g., \(t\)-tests and ANOVAs) is
employed; see \citet{applied-kuhn-2013} for details.

For classification problems, an area under the ROC curve (AUC) statistic
can be used to quantify predictor importance. The AUC statistic is
computed by using the predictor \(x\) as input to the ROC curve. If
\(x\) can reasonably separate the classes of \(Y\), that is a clear
indicator that \(x\) is an important predictor (in terms of class
separation) and this is captured in the corresponding AUC statistic. For
problems with more than two classes, extensions of the ROC curve or a
one-vs-all approach can be used.

\subsection{Partial dependence plots}

\citet{harrison-1978-hedonic} analyzed a data set containing suburban
Boston housing data from the 1970 census. They sought a housing value
equation using an assortment of features; see Table IV of
\citet{harrison-1978-hedonic} for a description of each variable. The
usual regression assumptions, such as normality, linearity, and constant
variance, were clearly violated, but through an exhausting series of
transformations, significance testing, and grid searches, they were able
to build a model which fit the data reasonably well (\(R^2 = 0.81\)).
Their prediction equation is given in Equation \eqref{eqn:boston}. This
equation makes interpreting the model easier. For example, the average
number of rooms per dwelling (\(RM\)) is included in the model as a
quadratic term with a positive coefficient. This means that there is a
monotonic increasing relationship between \(RM\) and the predicted
median home value, but larger values of \(RM\) have a greater impact.

\begin{equation}
\label{eqn:boston}
\begin{aligned}
\widehat{\log\left(MV\right)} &= 9.76 + 0.0063 RM^2 + 8.98\times10^{-5} AGE - 0.19\log\left(DIS\right) + 0.096\log\left(RAD\right) \\
  & \quad - 4.20\times10^{-4} TAX - 0.031 PTRATIO + 0.36\left(B - 0.63\right)^2 - 0.37\log\left(LSTAT\right) \\
  & \quad - 0.012 CRIM + 8.03\times10^{-5} ZN + 2.41\times10^{-4} INDUS + 0.088 CHAS \\
  & \quad - 0.0064 NOX^2.
\end{aligned}
\end{equation}

However, classical regression and model building is rather ill-suited
for more contemporary big data sets, like the Ames housing data
described in \citet{ames-cock-2011} which has a total of 79 predictors
(and many more that can be created through feature engineering).
Fortunately, using modern computing power, many supervised learning
algorithms can fit such data sets in seconds, producing powerful, highly
accurate models. The downfall of many of these machine learning
algorithms, however, is decreased interpretability. For example, fitting
a well-tuned RF to the Boston housing data will likely produce a model
with more accurate predictions, but no interpretable prediction formula
such as the one in Equation \eqref{eqn:boston}.

To help understand the estimated functional relationship between each
predictor and the outcome of interest in a fitted model, we can
construct PDPs. PDPs are particularly effective at helping to explain
the output from ``black box'' models, such as RFs and SVMs. Not only do
PDPs visually convey the relationship between low cardinality subsets of
the feature set (usually 1-3) and the response (while accounting for the
average effect of the other predictors in the model), they can also be
used to rank and score the predictors in terms of their relative
influence on the predicted outcome, as will be demonstrated in this
paper.

Let \(\boldsymbol{x} = \left\{x_1, x_2, \dots, x_p\right\}\) represent
the predictors in a model whose prediction function is
\(\widehat{f}\left(\boldsymbol{x}\right)\). If we partition
\(\boldsymbol{x}\) into an interest set, \(\boldsymbol{z}_s\), and its
complement,
\(\boldsymbol{z}_c = \boldsymbol{x} \setminus \boldsymbol{z}_s\), then
the ``partial dependence'' of the response on \(\boldsymbol{z}_s\) is
defined as

\begin{equation}
\label{eqn:avg_fun}
  f_s\left(\boldsymbol{z}_s\right) = E_{\boldsymbol{z}_c}\left[\widehat{f}\left(\boldsymbol{z}_s, \boldsymbol{z}_c\right)\right] = \int \widehat{f}\left(\boldsymbol{z}_s, \boldsymbol{z}_c\right)p_{c}\left(\boldsymbol{z}_c\right)d\boldsymbol{z}_c,
\end{equation}

where \(p_{c}\left(\boldsymbol{z}_c\right)\) is the marginal probability
density of \(\boldsymbol{z}_c\):
\(p_{c}\left(\boldsymbol{z}_c\right) = \int p\left(\boldsymbol{x}\right)d\boldsymbol{z}_s\).
Equation \eqref{eqn:avg_fun} can be estimated from a set of training
data by

\begin{equation}
\label{eqn:pdf}
\bar{f}_s\left(\boldsymbol{z}_s\right) = \frac{1}{n}\sum_{i = 1}^n\widehat{f}\left(\boldsymbol{z}_s,\boldsymbol{z}_{i, c}\right),
\end{equation}

where \(\boldsymbol{z}_{i, c}\) \(\left(i = 1, 2, \dots, n\right)\) are
the values of \(\boldsymbol{z}_c\) that occur in the training sample;
that is, we average out the effects of all the other predictors in the
model.

Constructing a PDP \eqref{eqn:pdf} in practice is rather
straightforward. To simplify, let \(\boldsymbol{z}_s = x_1\) be the
predictor variable of interest with unique values
\(\left\{x_{11}, x_{12}, \dots, x_{1k}\right\}\). The partial dependence
of the response on \(x_1\) can be constructed as follows:

\begin{algorithm}
  \textbf{Input}: the unique predictor values $x_{11}, x_{12}, \dots, x_{1k}$;

  \textbf{Output}: the estimated partial dependence values $\bar{f}_1\left(x_{11}\right), \bar{f}_1\left(x_{12}\right), \dots, \bar{f}_1\left(x_{1k}\right)$. \BlankLine
  \For{$i \in \left\{1, 2, \dots, k\right\}$}{
    \BlankLine
    (1) copy the training data and replace the original values of $x_1$ with the constant $x_{1i}$;
    \BlankLine
    (2) compute the vector of predicted values from the modified copy of the training data;
    \BlankLine
    (3) compute the average prediction to obtain $\bar{f}_1\left(x_{1i}\right)$.\BlankLine}
    \BlankLine
    The PDP for $x_1$ is obtained by plotting the pairs $\left\{x_{1i}, \bar{f}_1\left(x_{1i}\right)\right\}$ for $i = 1, 2, \dotsc, k$. 
    \BlankLine
    \BlankLine
  \caption{A simple algorithm for constructing the partial dependence of the response on a single predictor $x_1$. \label{alg:pdp}}
\end{algorithm}

Algorithm \ref{alg:pdp} can be computationally expensive since it
involves \(k\) passes over the training records. Fortunately, it is
embarrassingly parallel and computing partial dependence functions for
each predictor can be done rather quickly on a machine with a multi-core
processor. For large data sets, it may be worthwhile to reduce the grid
size by using specific quantiles for each predictor, rather than all the
unique values. For example, the partial dependence function can be
approximated very quickly by using the deciles of the unique predictor
values. The exception is classification and regression trees based on
single-variable splits which can make use of the efficient weighted tree
traversal method described in \citet{friedman-2001-greedy}.

While PDPs are an invaluable tool in understanding the relationships
uncovered by complex nonparametric models, they can be misleading in the
presence of substantial interaction effects
\citep{goldstein-peeking-2015}. To overcome this issue,
\citeauthor{goldstein-peeking-2015} introduced the concept of individual
conditional expectation (ICE) curves. ICE curves display the estimated
relationship between the response and a predictor of interest for each
observation; in other words, skipping step 1 (c) in Algorithm
\ref{alg:pdp}. Consequently, the PDP for a predictor of interest can be
obtained by averaging the corresponding ICE curves across all
observations. Although ICE curves provide a refinement over traditional
PDPs in the presence of substantial interaction effects, in Section
\ref{sec:interaction}, we show how to use partial dependence functions
to evaluate the strength of potential interaction effects.

\subsection{The Ames housing data set}
\label{sec:ames}

\begin{figure}

{\centering \includegraphics{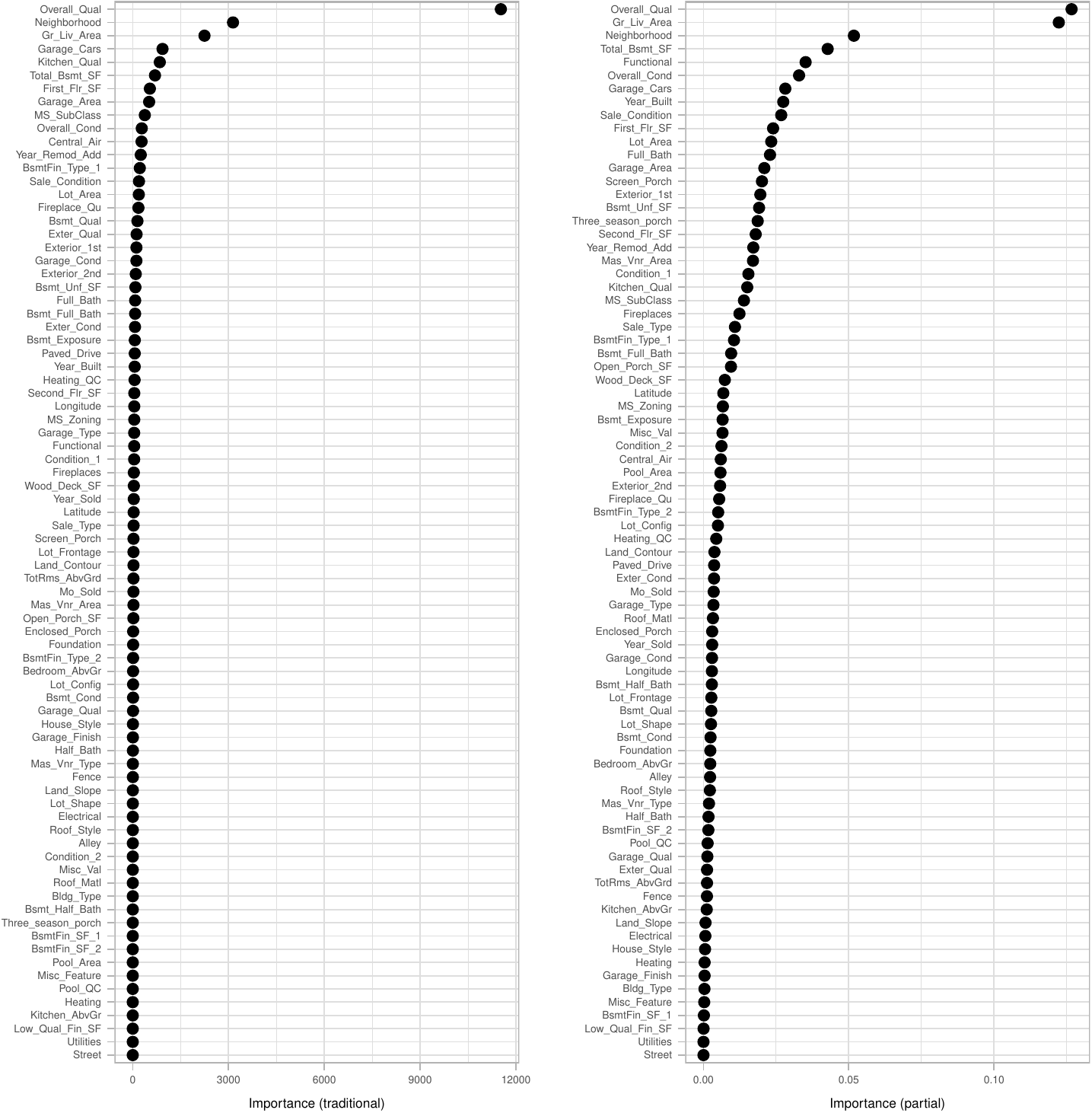} 

}

\caption{Variable importance scores for the Ames housing data. \textit{Left:} Traditional approach. \textit{Right:} Partial dependence-based approach.}\label{fig:ames-gbm-vip}
\end{figure}

For illustration, we will use the Ames housing data set---a modernized
and expanded version of the often cited Boston Housing data set. These
data are available in the \texttt{AmesHousing} package
\citep{AmesHousing-pkg}. Using the R package \texttt{h2o}
\citep{h2o-pkg}, we trained and tuned a GBM using 10-fold
cross-validation. The model fit is reasonable, with a cross-validated
(pseudo) \(R^2\) of 91.54\%. Like other tree-based ensembles, GBMs have
a natural way of defining variable importance which was described in
Section \ref{sec:model-based-vi}. The variable importance scores for
these data are displayed in the left side of Figure
\ref{fig:ames-gbm-vip}. This plot indicates that the overall quality of
the material and finish of the house (\texttt{Overall\_Qual}), physical
location within the Ames city limits (\texttt{Neighborhood}), and the
above grade (ground) living area (\texttt{Gr\_Liv\_Area}) are highly
associated with the logarithm of the sales price
(\texttt{Log\_Sale\_Price}). The variable importance scores also
indicate that pool quality (\texttt{Pool\_QC}), the number of kitchens
above grade (\texttt{Kitchen\_AbvGr}), and the low quality finished
square feet for all floors (\texttt{Low\_Qual\_Fin\_SF}) have little
association with \texttt{Log\_Sale\_Price}. (Note that the bottom two
features in Figure \ref{fig:ames-gbm-vip}---\texttt{Street} and
\texttt{Utilities}---have a variable importance score of exactly zero;
in other words, they were never used to partion the data at any point in
the GBM ensemble.)

\begin{figure}

{\centering \includegraphics{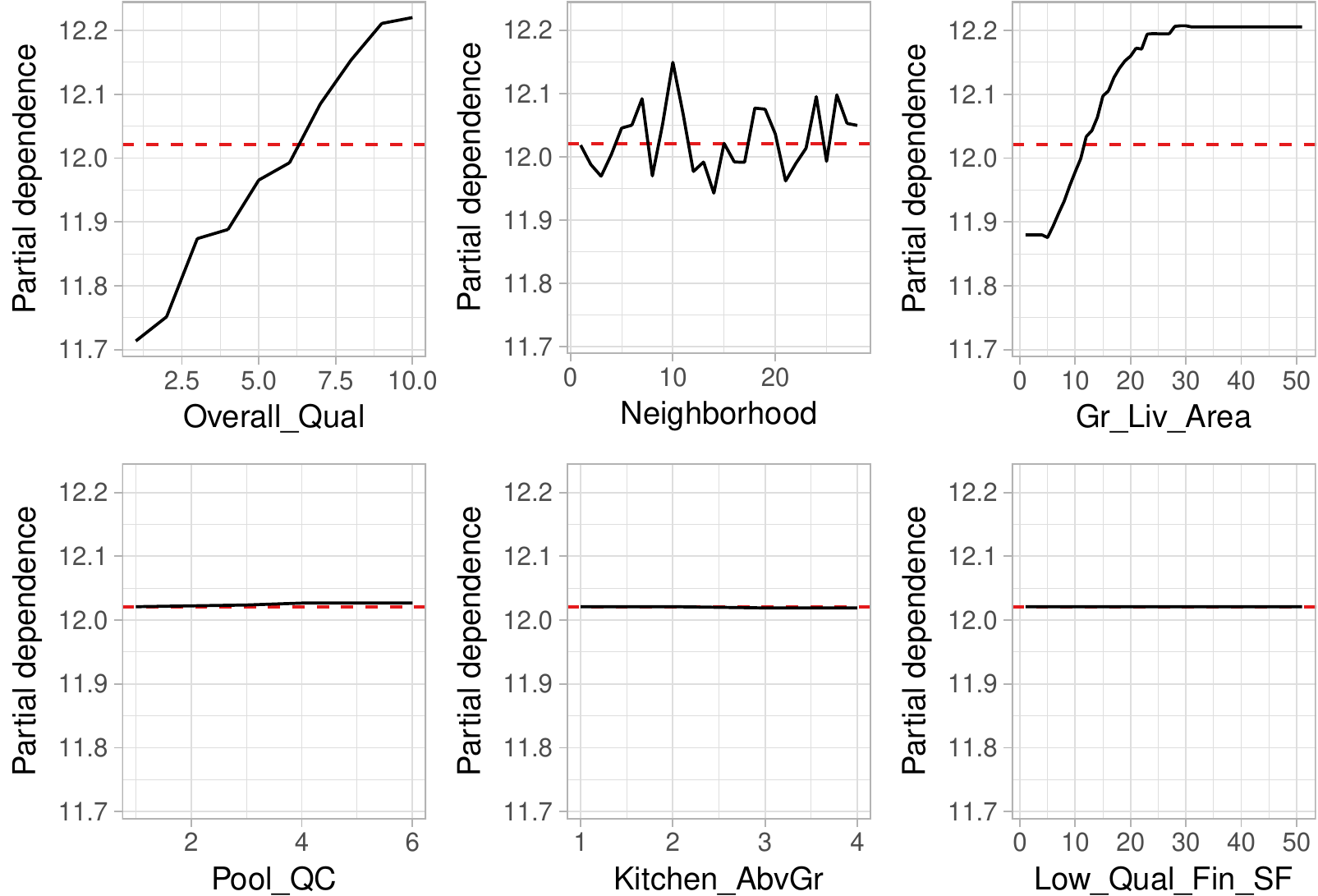} 

}

\caption{Partial dependence of log selling price on the three highest (top row) and lowest (bottom row) ranked predictors. The dashed red line in each plot represents the mean of the $N = 1460$ log selling prices}\label{fig:ames-gbm-pdps}
\end{figure}

The PDPs for these six variables are displayed in Figure
\ref{fig:ames-gbm-pdps}. These plots indicate that
\texttt{Overall\_Qual}, \texttt{Neighborhood}, and
\texttt{Gr\_Liv\_Area} have a strong nonlinear relationship with the
predicted outcome. For instance, it seems that \texttt{GrLivArea} has a
monotonically increasing relationship with \texttt{Log\_Sale\_Price}
until about 12 sq ft, after which the relationship flattens out. Notice
how the PDPs for \texttt{Pool\_QC}, \texttt{Kitchen\_AbvGr}, and
\texttt{Low\_Qual\_Fin\_SF} are relatively flat in comparison. It is
this notion of ``flatness'' which we will use as a basis to define our
variable importance measure.

\section{A partial dependence-based variable importance measure}
\label{sec:new}

The PDPs for \texttt{Pool\_QC}, \texttt{Kitchen\_AbvGr}, and
\texttt{Low\_Qual\_Fin\_SF} in Figure \ref{fig:ames-gbm-pdps} are
relatively flat, indicating that they do not have much influence on the
predicted value of \texttt{Log\_Sale\_Price}. In other words, the
partial dependence values \(\bar{f}_i\left(x_{ij}\right)\)
\(\left(j = 1, 2, \dots, k_i\right)\) display little variability. One
might conclude that any variable for which the PDP is ``flat'' is likely
to be less important than those predictors whose PDP varies across a
wider range of the response.

Our notion of variable importance is based on any measure of the
``flatness'' of the partial dependence function. In general, we define
\[
  i\left(x\right) = \digamma\left(\bar{f}_s\left(\boldsymbol{z}_s\right)\right),
\] where \(\digamma\left(\cdot\right)\) is any measure of the
``flatness'' of \(\bar{f}_s\left(\boldsymbol{z}_s\right)\). A simple and
effective measure to use is the sample standard deviation for continuous
predictors and the range statistic divided by four for factors with
\(K\) levels; the range divided by four provides an estimate of the
standard deviation for small to moderate sample sizes. Based on
Algorithm \ref{alg:pdp}, our importance measure for predictor \(x_1\) is
simply

\begin{equation}
\label{eqn:vi}
  i\left(x_1\right) =
  \begin{cases}
    \sqrt{\frac{1}{k - 1}\sum_{i = 1}^k\left[\bar{f}_1\left(x_{1i}\right) - \frac{1}{k}\sum_{i = 1}^k\bar{f}_1\left(x_{1i}\right)\right] ^ 2} & \quad \text{if } x_1 \text{ is continuous}\\
    \left[\max_i\left(\bar{f}_1\left(x_{1i}\right)\right) - \min_i\left(\bar{f}_1\left(x_{1i}\right)\right)\right] / 4 & \quad \text{if } x_1 \text{ is categorical}
  \end{cases}.
\end{equation}

Note that our variable importance metric relies on the fitted model;
hence, it is crucial to properly tune and train the model to attain the
best performance possible.

To illustrate, we applied Algorithm \ref{alg:pdp} to all of the
predictors in the Ames GBM model and computed \eqref{eqn:vi}. The
results are displayed in right side of Figure \ref{fig:ames-gbm-vip}. In
this case, our partial dependence-based algorithm matches closely with
the results from the GBM. In particular, Figure
\ref{fig:ames-gbm-vi-both} indicates that \texttt{Overall\_Qual},
\texttt{Neighborhood}, and \texttt{Gr\_Liv\_Area} are still the most
important variables in predicting \texttt{Log\_Sale\_Price}; though,
\texttt{Neighborhood} and \texttt{Gr\_Liv\_Area} have swapped places.

\subsection{Linear models}
\label{sec:linear}

As mentioned earlier, a natural choice for measuring the importance of
each term in a linear model is to use the absolute value of the
corresponding coefficient divided by its estimated standard error (i.e.,
the absolute value of the \(t\)-statistic). This turns out to be
equivalent to the partial dependence-based metric \eqref{eqn:vi} when
the predictors are independently and uniformly distributed over the same
range.

For example, suppose we have a linear model of the form

\begin{equation*}
  Y = \beta_0 + \beta_1 X_1 + \beta_2 X_2 + \epsilon,
\end{equation*}

where \(\beta_i\) (\(i = 1, 2\)) is a constant, \(X_1\) and \(X_2\) are
both independent \(\mathcal{U}\left(0, 1\right)\) random variables, and
\(\epsilon \sim \mathcal{N}\left(0, \sigma ^ 2\right)\). Since we know
the distribution of \(X_1\) and \(X_2\), we can easily find
\(f_1\left(X_1\right)\) and \(f_2\left(X_2\right)\). For instance,

\begin{equation*}
  f_1\left(X_1\right) = \int_0^1 E\left[Y | X_1, X_2\right] p\left(X_2\right) dX_2,
\end{equation*}

where \(p\left(X_2\right) = 1\). Simple calculus then leads to

\begin{equation*}
  f_1\left(X_1\right) = \beta_0 + \beta_2 / 2 + \beta_1 X_1 \quad and \quad f_2\left(X_2\right) = \beta_0 + \beta_1 / 2 + \beta_2 X_2..
\end{equation*}

Because \(E\left[Y | X_1, X_2\right] = f\left(X_1, X_2\right)\) is
additive, the true partial dependence functions are just simple linear
regressions in each predictor with their original coefficient and an
adjusted intercept. Taking the variance of each gives

\begin{equation*}
Var\left[f_1\left(X_1\right)\right] = \beta_1 ^ 2 / 12 \quad and \quad Var\left[f_2\left(X_2\right)\right] = \beta_2 ^ 2 / 12.
\end{equation*}

Hence, the standard deviations are just the absolute values of the
original coefficients (scaled by the same constant).

To illustrate, we simulated \(n = 1000\) observations from the following
linear model

\begin{equation*}
  Y = 1 + 3 X_1 - 5 X_2 + \epsilon,
\end{equation*}

where \(X_1\) and \(X_2\) are both independent
\(\mathcal{U}\left(0, 1\right)\) random variables, and
\(\epsilon \sim \mathcal{N}\left(0, 0.01 ^ 2\right)\). For this example,
we have

\begin{equation*}
  f_1\left(X_1\right) = -\frac{3}{2} + 3 X_1 \quad and \quad f_2\left(X_1\right) = \frac{5}{2} - 5 X_2.
\end{equation*}

\begin{figure}

{\centering \includegraphics[width=1.0\linewidth]{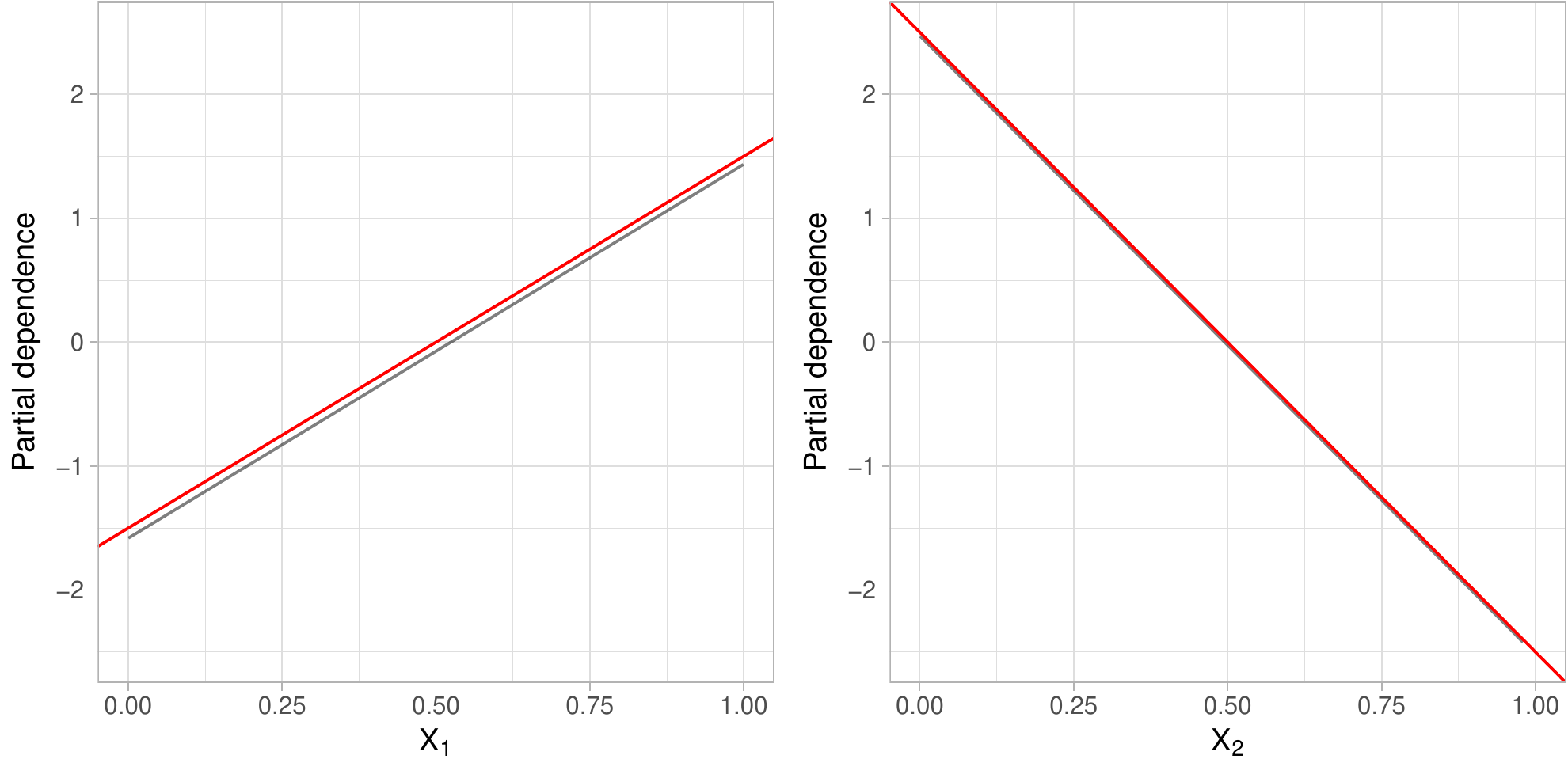} 

}

\caption{Estimated (black) and true (red) partial dependence functions from a linear model with two predictors.}\label{fig:lm-pdps}
\end{figure}

These are plotted as red lines in Figure \ref{fig:lm-pdps}.
Additionally, the black lines in Figure \ref{fig:lm-pdps} correspond to
the estimated partial dependence functions using Algorithm
\ref{alg:pdp}.

Based on these plots, \(X_2\) is more influential than \(X_1\). Taking
the absolute value of the ratio of the slopes in \(f_2\left(X_2\right)\)
and \(f_1\left(X_1\right)\) gives \(5 / 3 \approx 1.67\). In other
words, \(X_2\) is roughly 1.67 times more influential on \(\widehat{Y}\)
than \(X_1\). Using the partial-dependence-based variable importance
metric, we obtain \(i\left(X_1\right) = 1.4828203\) and
\(i\left(X_2\right) = 0.8961719\) which gives the ratio
\(i\left(X_2\right) / i\left(X_1\right) \approx 1.65\). In fact, we can
compute the ratio of the true variances of \(f_1\left(X_1\right)\) and
\(f_2\left(X_1\right)\):

\begin{equation*}
Var\left[f_2\left(X_2\right)\right] / Var\left[f_1\left(X_1\right)\right] = \left(5^2 / 12\right) / \left(3^2 / 12\right) = \left(5 / 3\right)^2.
\end{equation*}

Taking the square root gives \(5 / 3 \approx 1.67\).

Using the absolute value of the \(t\)-statistic becomes less useful in
linear models when, for example, a predictor appears in multiple terms
(e.g., interaction effects and polynomial terms). The partial dependence
approach, on the other hand, does not suffer from such drawbacks.

\subsection{Detecting interaction effects}
\label{sec:interaction}

As it turns out, our partial dependence-based variable importance
measure \eqref{eqn:vi} can also be used to quantify the strength of
potential interaction effects. Let \(\i\left(x_i, x_j\right)\)
\(\left(i \ne j\right)\) be the standard deviation of the joint partial
dependence values \(\bar{f}_{ij}\left(x_{ii'}, x_{jj'}\right)\) for
\(i' = 1, 2, \dots, k_i\) and \(j' = 1, 2, \dots, k_j\). Essentially, a
weak interaction effect of \(x_i\) and \(x_j\) on \(Y\) would suggest
that \(i\left(x_i, x_j\right)\) has little variation when either \(x_i\)
or \(x_j\) is held constant while the other varies.

Let \(\boldsymbol{z}_s = \left(x_i, x_j\right)\), \(i \neq j\), be any
two predictors in the feature space \(\boldsymbol{x}\). Construct the
partial dependence function \(\bar{f}_s\left(x_i, x_j\right)\) and
compute \(i\left(x_i\right)\) for each unique value of \(x_j\), denoted
\(\i\left(x_i | x_j\right)\), and take the standard deviation of the
resulting importance scores. The same can be done for \(x_j\) and the
results are averaged together. Large values (relative to each other)
would be indicative of possible interaction effects.

\section{Friedman's regression problem}

To further illustrate, we will use one of the regression problems
described in \citet{multivariate-friedman-1991} and
\citet{bagging-breiman-1996}. The feature space consists of ten
independent \(\mathcal{U}\left(0, 1\right)\) random variables; however,
only five out of these ten actually appear in the true model. The
response is related to the features according to the formula

\begin{equation*}
Y = 10 \sin\left(\pi x_1 x_2\right) + 20 \left(x_3 - 0.5\right) ^ 2 + 10 x_4 + 5 x_5 + \epsilon,
\end{equation*}

where \(\epsilon \sim \mathcal{N}\left(0, \sigma^2\right)\). Using the R
package \texttt{nnet} \citep{venables-modern-2002}, we fit a NN with one
hidden layer containing eight units and a weight decay of 0.01 (these
parameters were chosen using 5-fold cross-validation) to 500
observations simulated from the above model with \(\sigma = 1\). The
cross-validated \(R^2\) value was 0.94.

\begin{figure}

{\centering \includegraphics[width=1.0\linewidth]{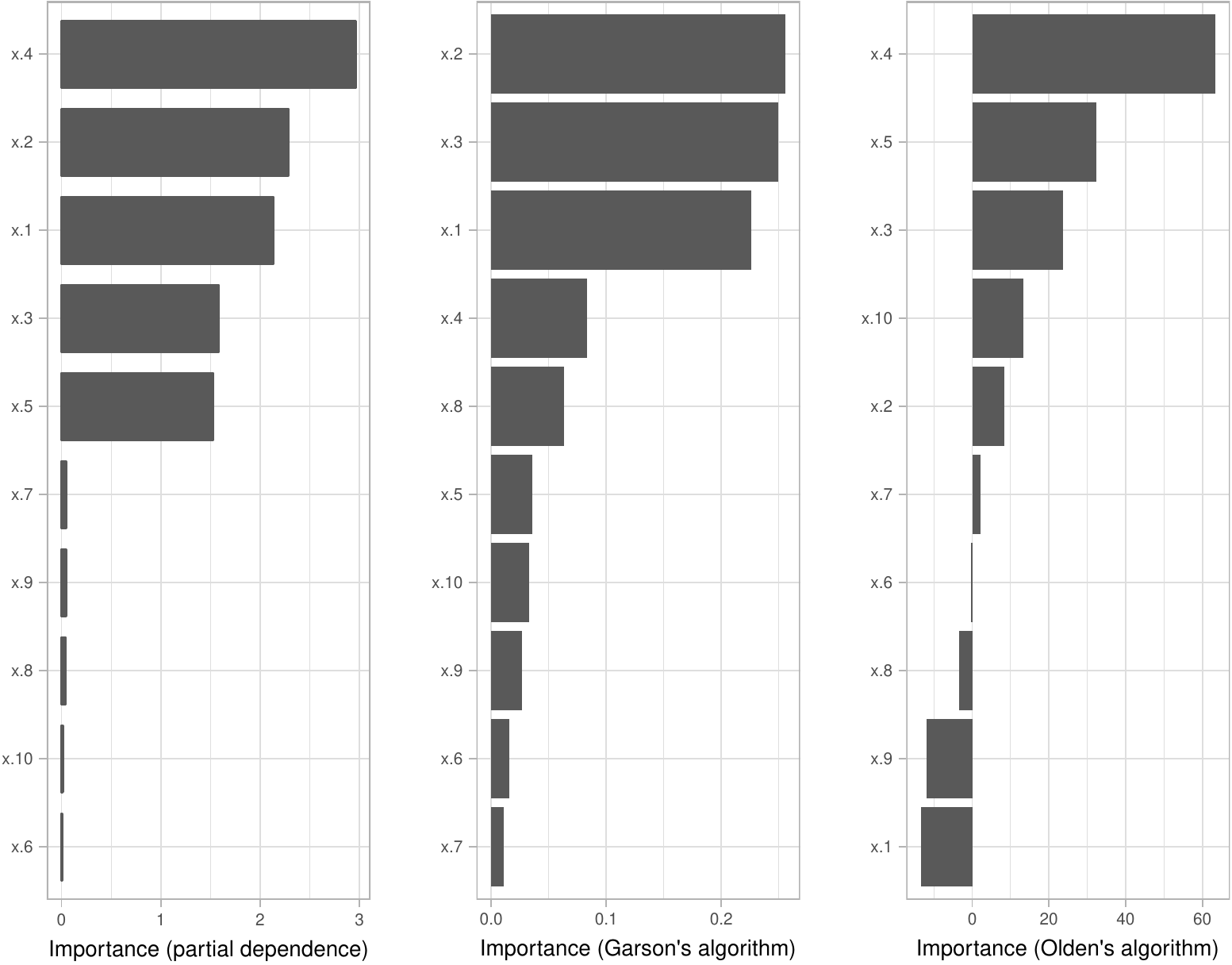} 

}

\caption{Variable importance plots for the NN fit to the Friedman regression data. \textit{Left:} partial dependence-based method. \textit{Middle:} Garson's method. \textit{Right:} Olden's method.}\label{fig:network-vip}
\end{figure}

Variable importance plots are displayed in Figure \ref{fig:network-vip}.
Notice how the Garson and Olden algorithms incorrectly label some of the
features not in the true model as ``important''. For example, the Garson
algorithm incorrectly labels \(x_8\) (which is not included in the true
model) as more important than \(x_5\) (which is in the true model).
Similarly, Oden's method incorrectly labels \(x_{10}\) as being more
important than \(x_2\). Our method, on the other hand, clearly labels
all five of the predictors in the true model as the most important
features in the fitted NN.

\begin{figure}

{\centering \includegraphics[width=1.0\linewidth]{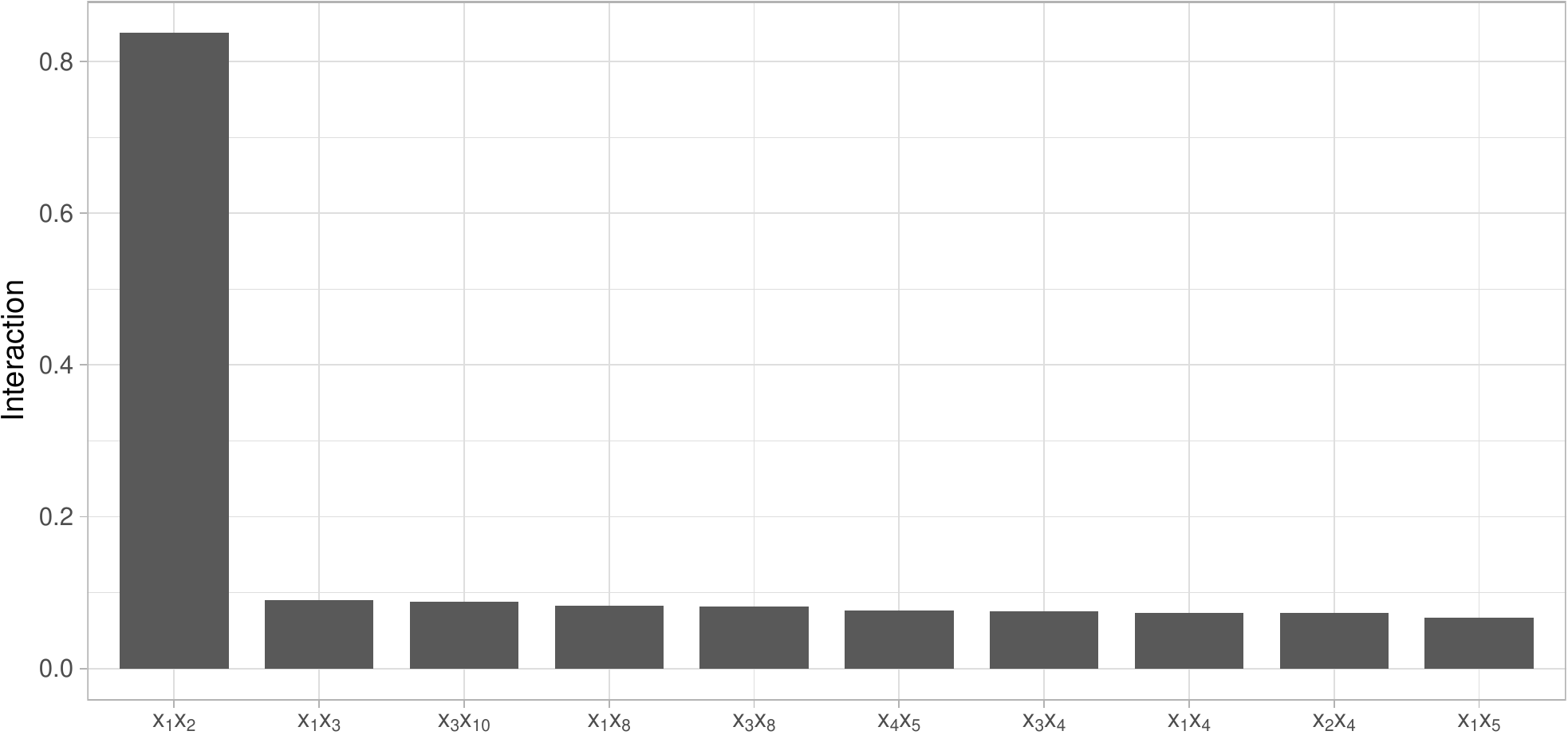} 

}

\caption{Variable importance-based interaction statistics from the NN fit to the Friedman regression data set.}\label{fig:network-int}
\end{figure}

We also constructed the partial dependence functions for all pairwise
interactions and computed the interaction statistic discussed in Section
\ref{sec:interaction}. The top ten interaction statistics are displayed
in Figure \ref{fig:network-int}. There is a clear indication of an
interaction effect between features \(x_1\) and \(x_2\), the only
interaction effect present in the true model.

In fact, since we know the distributions of the predictors in the true
model, we can work out the true partial dependence functions. For
example, for the pairs \(\left(x_1, x_2\right)\) and
\(\left(x_1, x_4\right)\), we have

\begin{equation*}
f\left(x_1, x_2\right) = 10 \sin \left(\pi x_1 x_2\right) + 55 / 6,
\end{equation*}

and

\begin{equation*}
f\left(x_1, x_4\right) = \frac{5 \pi x_1 \left(12 x_4 + 5\right) - 12 \cos \left(\pi x_1\right) + 12}{6 \pi x_1}.
\end{equation*}

\begin{figure}

{\centering \includegraphics[width=1.0\linewidth]{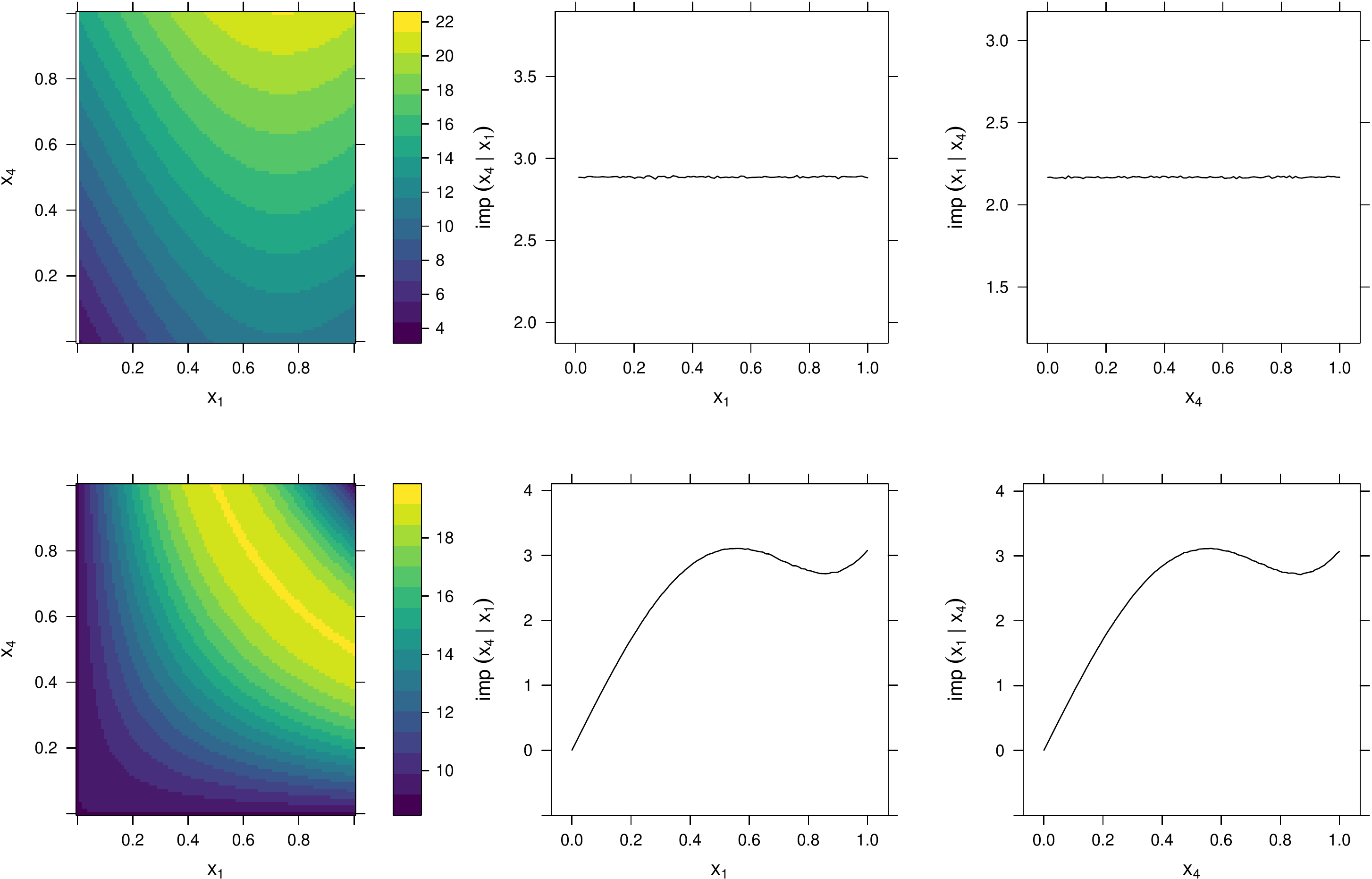} 

}

\caption{Results from a small Monte Carlo simulation on the interaction effects between $x_1$ and $x_2$ (top row), and $x_1$ and $x_4$ (bottom row).}\label{fig:interaction-simulation}
\end{figure}

Next, we simulated the standard deviation of \(f\left(x_1, x_2\right)\)
for a wide range of fixed values of \(x_2\); this is what
\(i\left(x_1 | x_2\right)\) is trying to estimate. The results from
doing this for both predictors in each model are displayed in
Figure\textasciitilde{}\ref{fig:interaction-simulation}. The top row of
Figure\textasciitilde{}\ref{fig:interaction-simulation} illustrates that
the importance of \(x_1\) (i.e., the strength of its relationship to the
predicted outcome) heavily depends on the value of \(x_2\) and vice
versa (i.e., an interaction effect between \(x_1\) and \(x_2\)). In the
bottom row, on the other hand, we see that the importance of \(x_1\)
does not depend on the value of \(x_4\) and vice versa (i.e., no
interaction effect between \(x_1\) and \(x_4\)).

\subsection{Friedman's $H$-statistic}

An alternative measure for the strength of interaction effects is known
as Friedman's \(H\)-statistic \citep{friedman-2008-predictive}.
Coincidentally, this method is also based on the estimated partial
dependence functions of the corresponding predictors, but uses a
different approach.

\begin{figure}

{\centering \includegraphics[width=1.0\linewidth]{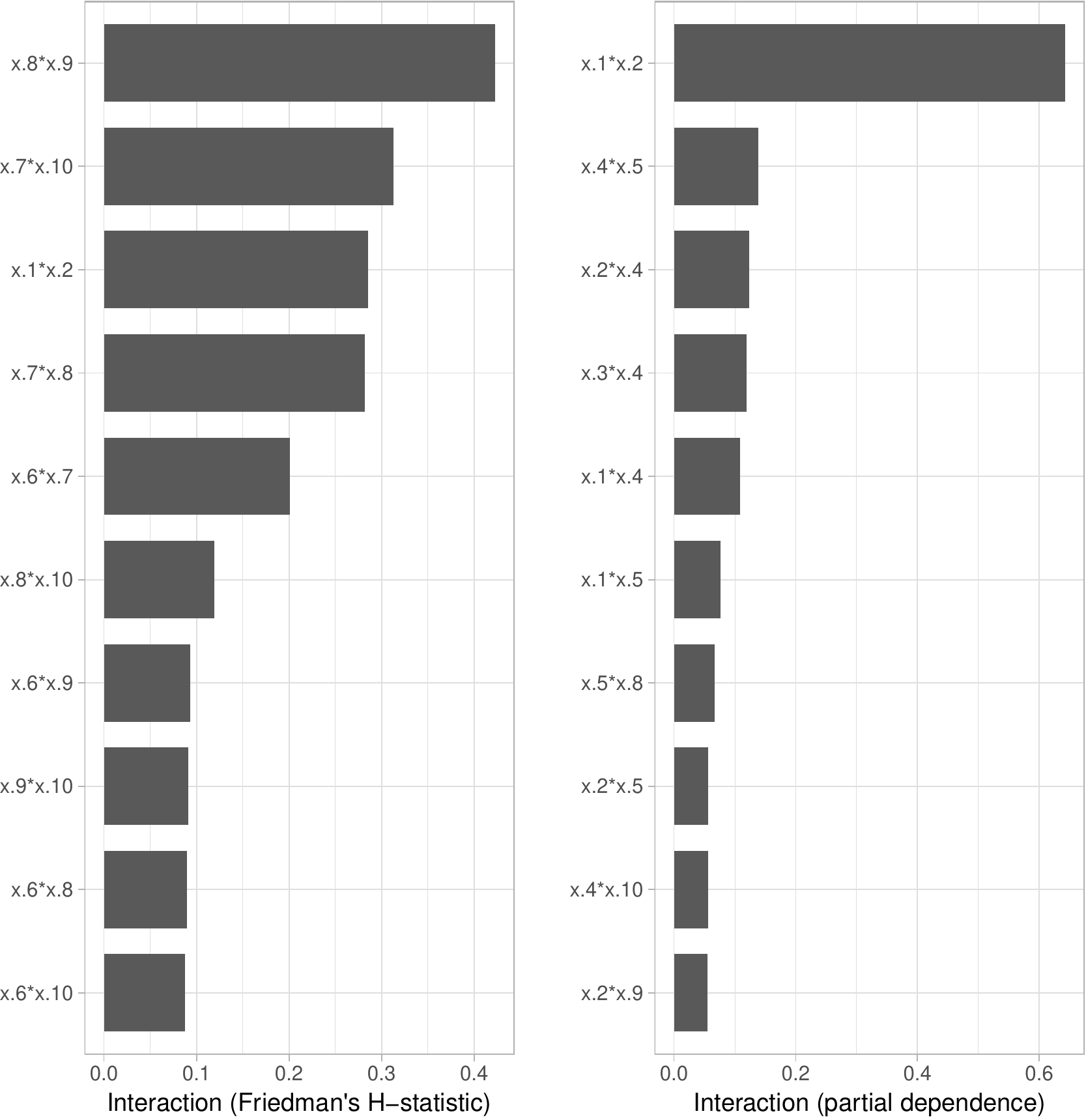} 

}

\caption{Interaction statistics for the GBM model fit to the Friedman regression data. \textit{Left:} Friedman's $H$-statistic. \textit{Right:} Our variable importance-based interaction statistic.}\label{fig:gbm-int}
\end{figure}

For comparison, we fit a GBM to the Friedman regression data from the
previous section. The parameters were chosen using 5-fold
cross-validation. We used the R package \texttt{gbm} \citep{gbm-pkg}
which has built-in support for computing Friedman's \(H\)-statistic for
any combination of predictors. The results are displayed in Figure
\ref{fig:gbm-int}. To our surprise, the \(H\)-statistic did not seem to
catch the true interaction between \(x_1\) and \(x_2\). Instead, the
\(H\)-statistic ranked the pairs \(\left(x_8, x_9\right)\) and
\(\left(x_7, x_{10}\right)\) as having the strongest interaction
effects, even though these predictors do not appear in the true model.
Our variable importance-based interaction statistic, on the other hand,
clearly suggests the pair \(\left(x_1, x_2\right)\) as having the
strongest interaction effect.

\section{Application to model stacking}
\label{sec:ensemble}

In the Ames housing example, we used a GBM to illustrate the idea of
using the partial dependence function to quantify the importance of each
predictor on the predicted outcome (in this case,
\texttt{Log\_Sale\_Price}). While the GBM model achieved a decent
cross-validated \(R^2\) of 91.54\%, better predictive performance can
often be attained by creating an \emph{ensemble} of learning algorithms.
While GBMs are themselves ensembles, we can use a method called
\emph{stacking} to combine the GBM with other cutting edge learning
algorithms to form a \emph{super learner} \citep{stacked-wolpert-1992}.
Such stacked ensembles tend to outperform any of the individual base
learners (e.g., a single RF or GBM) and have been shown to represent an
asymptotically optimal system for learning \citep{super-laan-2003}.

For the Ames data set, in addition to the GBM, we trained and tuned a RF
using 10-fold cross-validation. The cross-validated predicted values
from each of these models were combined to form a \(n \times 2\) matrix,
where \(n = 1460\) is the number of training records. This matrix,
together with the observed values of \texttt{Log\_Sale\_Price}, formed
the ``level-one'' data. Next, we trained a \emph{metalearning}
algorithm---in this case a GLM---on the level-one data to create a
stacked ensemble. To generate new predictions, we first generate
predictions from the RF and GBM learners, then feed those into the GLM
metalearner to generate the ensemble prediction.

\begin{figure}

{\centering \includegraphics[width=1.0\linewidth]{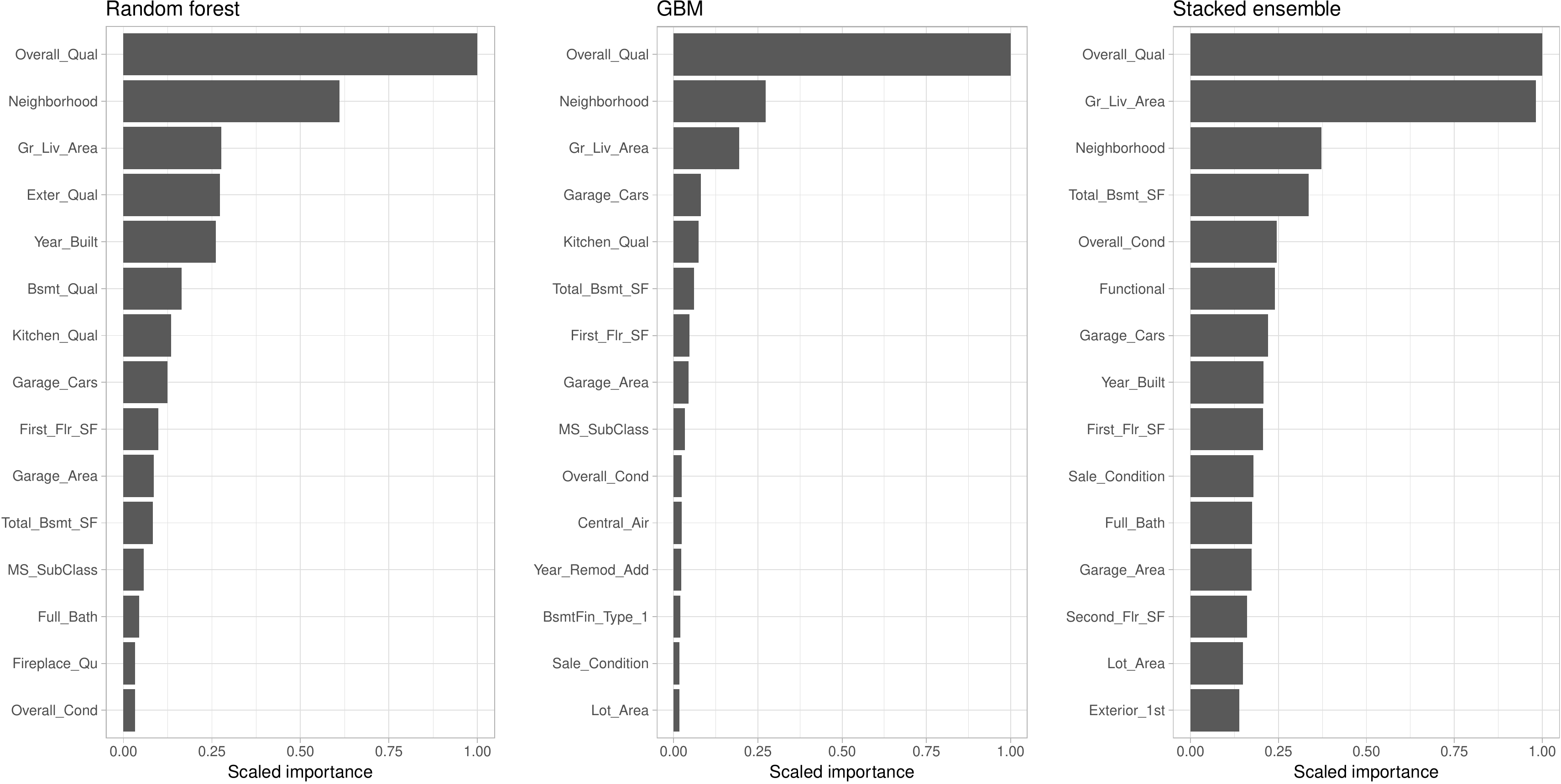} 

}

\caption{Variable importance plots for the Ames data set. \textit{Left:} RF. \textit{Middle:} GBM. \textit{Right:} Stacked ensemble (i.e., super learner).}\label{fig:ames-ensemble-vip}
\end{figure}

Even though the base learners in this example have the built-in
capability to compute variable importance scores, there is no way of
constructing them from the super learner. However, since we can generate
predictions from the super learner, we can easily construct partial
dependence functions for each predictor and use Equation \eqref{eqn:vi}.
The left and middle plots in Figure \ref{fig:ames-ensemble-vip} display
the top 15 variable importance scores from the individual base learners.
The right plot displays the top 15 variable importance scores
constructed using Equation \eqref{eqn:vi} for the combined super
learner. All models seem to agree on the top three predictors of
\texttt{Log\_Sale\_Price}; namely, \texttt{Overall\_Qual},
\texttt{Neighborhood}, and \texttt{Gr\_Liv\_Area}.

\section{Application to automatic machine learning}
\label{sec:automl}

The data science field has seen an explosion in interest over the last
decade. However, the supply of data scientists and machine learning
experts has not caught up with the demand. Consequently, many
organizations are turning to automated machine learning (AutoML)
approaches to predictive modelling. AutoML has been a topic of
increasing interest over the last couple of years and open source
implementations---like those in H2O and auto-sklearn
\citep{auto-sklearn}---have made it a simple and viable options for real
supervised learning problems.

Current AutoML algorithms leverage recent advantages in Bayesian
optimization, meta-learning, and ensemble construction (i.e., model
stacking). The benefit, as compared to the stacked ensemble formed in
Section \ref{sec:ensemble}, is that AutoML frees the analyst from having
to do algorithm selection (e.g., ``Do I fit an RF or GBM to my data?'')
and hyperparameter tuning (e.g., how many hidden layers to use in a
DNN). While AutoML has great practical potential, it does not
automatically provide any useful interpretations---AutoML is
\textbf{not} automated data science \citep{mayo_2017}. For instance,
which variables are the most important in making accurate predictions?
How do these variables functionally related to the outcome of interest?
Fortunately, our approach can still be used to answer these two
questions simultaneously.

To illustrate, we used the R implementation of H2O's AutoML algorithm to
model the airfoil self-noise data set \citep{uciml} which are available
from the University of California Machine Learning Repository:
\url{https://archive.ics.uci.edu/ml/data sets/airfoil+self-noise}. These
data are from a NASA experiment investigating different size NACA 0012
airfoils at various wind tunnel speeds and angles of attack. The
objective is to accurately predict the scaled sound pressure level (dB)
(\texttt{scaled\_sound\_pressure\_level}) using frequency (Hz)
(\texttt{frequency}), angle of attack (degrees)
(\texttt{angle\_of\_attack}), chord length (m) (\texttt{chord\_length}),
free-stream velocity (m/s) (\texttt{free\_stream\_velocity}), and
suction side displacement thickness (m)
(\texttt{suction\_side\_displacement\_thickness}). H2O's current AutoML
implementation trains and cross-validates an RF, an extremely-randomized
forest (XRF) \citep{extremely-geurts-2006}, a random grid of GBMs, a
random grid of DNNs, and then trains a stacked ensemble using the
approach outlined in Section \ref{sec:ensemble}. We used 10-fold
cross-validation and RMSE for the validation metric. Due to hardware
constraints, we used a max run time of 15 minutes (the default is one
hour). The final model consisted of one DNN, a random grid of 37 GBMs,
one GLM, one RF, one XRT, and two stacked ensembles. The final stacked
ensemble achieved a 10-fold cross-validated RMSE and R-squared of 1.43
and 95.68, respectively.

\begin{figure}

{\centering \includegraphics[width=1.0\linewidth]{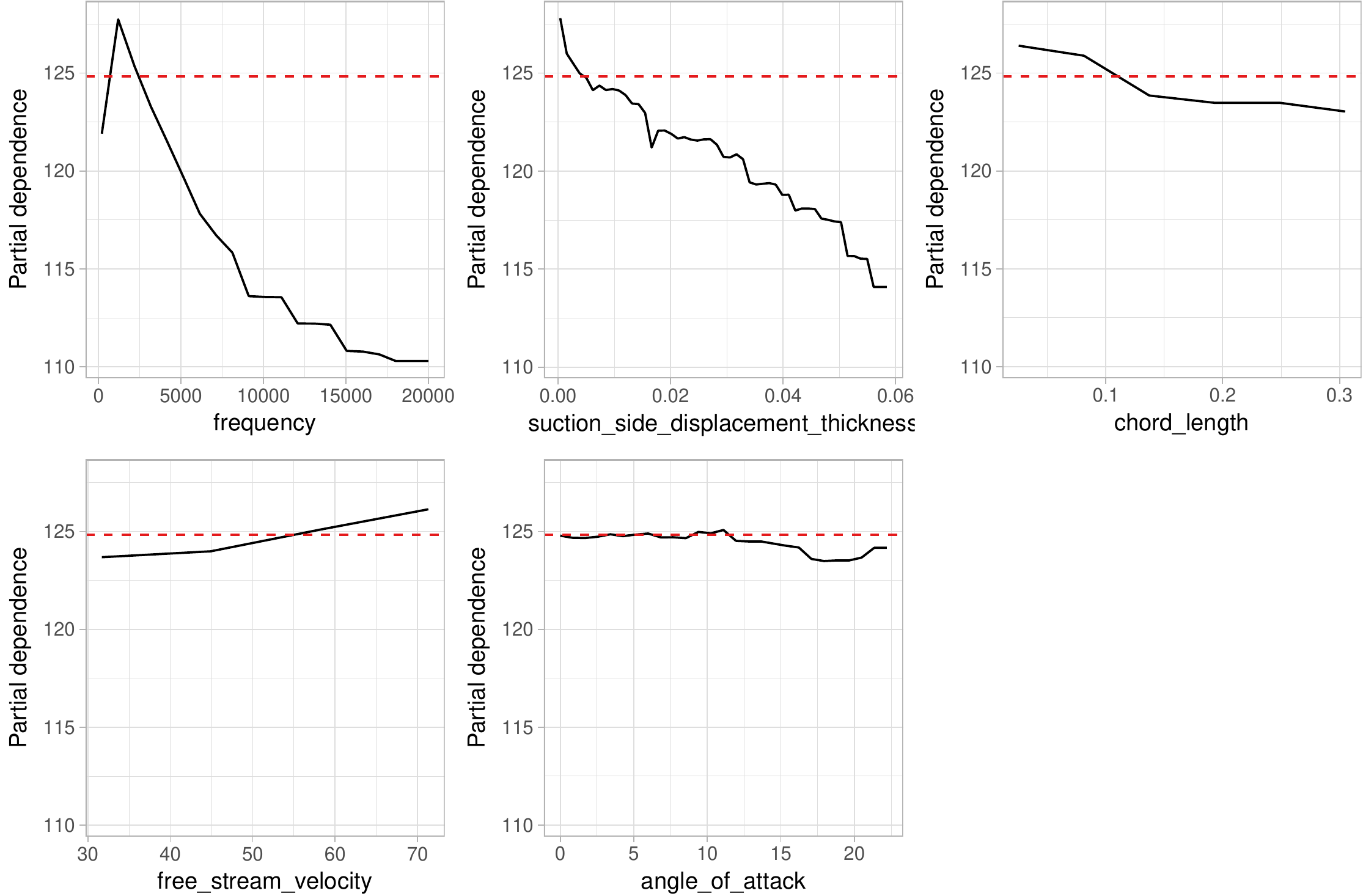} 

}

\caption{PDPs for the five features in the airfoil self-noise data set based on an AutoML algorithm consisting of DNNs, GBMs, GLMs, RFs, XRTs, and stacked ensembles. The dashed red line in each plot represents the mean scaled sound pressure level (dB) in the training data.}\label{fig:airfoil-aml-pdps}
\end{figure}

Since predictions can be obtained from the automated stacked ensemble,
we can easily apply Algorithm \ref{alg:pdp} and construct PDPs for all
the features; these are displayed in Figure \ref{fig:airfoil-aml-pdps}.
It seems frequency (Hz) and suction side displacement thickness (m) have
a strong monotonically decreasing relationship with scaled sound
pressure level (dB). They also appear to be more influential than the
other three. Using Equation \eqref{eqn:vi}, we computed the variable
importance of each of the five predictors which are displayed in Table
\ref{tab:airfoil-aml-vi}.

\begin{table}[!htb]
  \centering
  \begin{tabular}{lr}
    \toprule
    Variable                                & Importance \\
    \midrule
    Angle of attack (degrees)               & 0.4888482  \\
    Free stream velocity (m/s)              & 1.1158569  \\
    Chord length (m)                        & 1.4168613  \\
    Suction side displacement thickness (m) & 3.3389105  \\
    Frequency (Hz)                          & 5.4821362  \\
    \bottomrule
  \end{tabular}
  \caption{Partial dependence-based variable importance scores for the five predictors in the airfoil self-noise data set based on an AutoML algorithm consisting of DNNs, GBMs, GLMs, RFs, XRTs, and stacked ensembles. \label{tab:airfoil-aml-vi}}
\end{table}

\section{Discussion}
\label{sec:conc}

We have discussed a new variable importance measure that is (i) suitable
for use with any supervised learning algorithm, provided new predictions
can be obtained, (ii) model-based and takes into account the effect of
all the features in the model, (iii) consistent and has the same
interpretation regardless of the learning algorithm employed, and (iv)
has the potential to help identify possible interaction effects. Since
our algorithm is model-based it requires that the model be properly
trained and tuned to achieve optimum performance. While this new
approach appears to have high utility, more research is needed to
determine where its deficiencies may lie. For example, outliers in the
feature space can cause abnormally large fluctuations in the partial
dependence values \(\bar{f}\left(x_{i}\right)\)
\(\left(i = 1, 2, \dots, k\right)\). Therefore, it may be advantageous
to use more robust measures of spread to describe the variability in the
estimated partial dependence values; a reasonable choice would be the
median absolute deviation which has a finite sample breakdown point of
\(\left\lfloor{k / 2}\right\rfloor / k\). It is also possible to replace
the mean in step (3) of Algorithm \ref{alg:pdp} with a more robust
estimate such as the median or trimmed mean. Another drawback is the
computational burden imposed by Algorithm \ref{alg:pdp} on large data
sets, but this can be mitigated using the methods discussed in
\citet{pdp-greenwell-2017}.

All the examples in this article were produced using R version 3.4.0
\citep{R}; a software environment for statistical computing. With the
exception of Figure \ref{fig:interaction-simulation}, all graphics were
produced using the R package \texttt{ggplot2} \citep{pkg-ggplot2};
Figure \ref{fig:interaction-simulation} was produced using the R package
\texttt{lattice} \citep{pkg-lattice}.

\bigskip

\begin{center}
{\large\bf SUPPLEMENTARY MATERIAL}
\end{center}

\begin{description}

\item[R package:] The R package `vip`, hosted on GitHub at \url{https://github.com/AFIT-R/vip}, contains functions for computing variable importance scores and constructing variable importance plots for various types of fitted models in R using the partial dependence-based approach discussed in this paper.

\item[R code:] The R script `vip-2018.R` contains the R code to reproduce all of the results and figures in this paper.

\end{description}

\bibliographystyle{agsm}
\bibliography{bibliography.bib}

\end{document}